\documentclass[sigconf]{acmart}

\begin{document}


\copyrightyear{2017} 
\acmYear{2017} 
\setcopyright{acmcopyright}
\acmConference{DSMM'17}{May 14, 2017}{Chicago, IL, USA}\acmPrice{15.00}\acmDOI{http://dx.doi.org/10.1145/3077240.3077249}
\acmISBN{978-1-4503-5031-0/17/05}




%

\title{Predicting Role Relevance with Minimal Domain Expertise in a Financial Domain}

\author{Mayank Kejriwal}
\affiliation{%
  \institution{Information Sciences Institute}
  \streetaddress{USC Viterbi School of Engineering}
  \city{Marina Del Rey} 
  \state{CA} 
  \postcode{90292}
}
\email{kejriwal@isi.edu}



\begin{abstract}
Word embeddings have made enormous inroads in recent years in a wide variety of text mining applications. In this paper, we explore a word embedding-based architecture for predicting the relevance of a role between two financial entities within the context of natural language sentences. In this extended abstract, we propose a pooled approach that uses a collection of sentences to train word embeddings using the skip-gram word2vec architecture. We use the word embeddings to obtain context vectors that are assigned one or more labels based on manual annotations. We train a machine learning classifier using the labeled context vectors, and use the trained classifier to predict contextual role relevance on test data. Our approach serves as a good minimal-expertise baseline for the task as it is simple and intuitive, uses open-source modules, requires little feature crafting effort and performs well across roles.
\end{abstract}
\maketitle

%
%


%
%

%
%


\section*{Methods and Results}
{\bf Problem Statement and Preliminaries:} Given a triple $t =$ \\$(f_h,r,f_t)$ that consists of a \emph{head} financial entity $f_h$, a \emph{tail} financial entity $f_t$, and a \emph{role} $r$ , typically from a closed domain-specific universe $R$ of roles, \emph{role relevance prediction} is defined as the problem of assigning a relevance score in $[0,1]$ to the triple in a given context $c$. In the FEIII 2017 challenge, $c$ is just a set of three sentences in financial document filings \emph{by} the head entity, with the tail entity \emph{extracted} from the sentences. For the preliminary results in this paper, we use the labeled data provided by the challenge organizers for predicting role relevance.

Role relevance is not unlike an ad-hoc Information Retrieval (IR) problem in that the \emph{relevance} function is domain-specific and ad-hoc \cite{adhocIR}. This is indicated by the fact that, in some instances, manual annotators disagree on relevance. One reason could be that the two entities are subjectively relevant (in a generic sense), but not with respect to the stated role or context in the triple. 

However, role relevance is also different from ad-hoc IR because, unlike traditional IR, where a query is usually phrasal and a list of ranked documents is retrieved and evaluated on metrics like Normalized Discounted Cumulative Gain (NDCG), the query in role relevance is simply a role $r_q$, and the goal is to rank available triples in a financial knowledge graph with the caveat that (1) the role $r$ in the triple is exactly $r_q$ for a non-zero relevance score to be assigned to the triple, (2) the triple must be scored `contextually'. 

In a machine learning framework, one can approach this problem by using labeled training data to train a classifier for each role. In order to do this effectively, one also has to devise `feature crafting functions' that convert each triple and context to a feature vector. Feature crafting often involves trial and error, and manual labor, usually in increasing proportion to domain specificity. For example, one could search for the presence of certain `role-relevant' keywords in the context as features that indicate high role relevance.

{\bf Approach: }In contrast with feature crafting approaches, we propose an approach that uses skip-gram neural networks to convert contexts into \emph{role-independent}, \emph{low-dimensional} feature vectors to leverage data in a pooled fashion across roles, and then trains a \emph{role-specific classifier} like a random forest to learn the nuances of each role. 

The overall approach is described as follows. First, we collect available context sentences, whether labeled (either as relevant, highly relevant or irrelevant with respect to any given role) or unlabeled, and build a `corpus' where each sentence is akin to a document. We train a word embedding model using \emph{skip-gram word2vec} \cite{word2vec}, and with only 30 dimensions in the latent space. The low dimensionality is to prevent the vectors from being too unstable, as the domain-specific corpus is not large enough (i.e. does not have enough unique words) to usefully accommodate more parameters. Each word vector thus obtained is normalized by the model to lie on the 30-dimensional unit hypersphere. The dimensionality was set heuristically; we leave for future work to evaluate sensitivity of our results to this hyperparameter.

Once a word embedding model has been trained, we can generate feature vectors for each set of context sentences in a triple in several different ways. For the initial system we built, described herein, we consider a normalized bag-of-words average: we add the corresponding vectors of all words occurring in the context sentences, without regard for word order, and then (l2-)normalize the summed vector so that it lies on the same hypersphere as each word vector. Note that the number of times a word occurs in the context sentences \emph{does} matter as it will get included in the sum each time it occurs. In that sense, the summed `context feature vector' (CFV) thus obtained is like a low-dimensional, continuous version of a document vector obtained using a classic measure like tf-idf.   

We can now approach the problem from a traditional (binary classification) machine learning standpoint by training a different classifier for each \emph{role}. For our experiments, we use a random forest classifier, often known to work extremely well, and for simplicity, all contextual triples marked as `highly relevant' or `relevant' are assigned a positive-class label of 1.0 while `irrelevant' triples are assigned a negative-class label of 0.0. The CFV is the feature vector assigned to the triple. At test time, we compute the CFV for the context sentences accompanying the contextual test triple, and apply the classifier corresponding to the role in the triple. The score output by the classifier, always a real value between 0.0 and 1.0 is \emph{interpreted} as the role relevance probability of the triple. These scores can then be used to compute metrics such as precision, recall and NDCG, after ranking the triples in decreasing order of score.        
    
{\bf Advantages:} The proposed approach has some important advantages. Perhaps the most important advantage is that it does not require a human to devise `clever' features or an arduous trial-and-error process of labor-intensive feature experimentation. A second important advantage is that, due to using machine learning on two fronts (feature crafting, as well as relevance prediction), any additional data (whether labeled or unlabeled) can be leveraged to improve performance. For example, in the FEIII 2017 role relevance prediction challenge, a much larger (than the labeled training set) `working' set of unlabeled contextual triples\footnote{By a contextual triple, we  mean a triple that is accompanied by a set (usually three) of context sentences.} was released by challenge organizers. Clearly, the availability of this extra information, even without labels, presents an opportunity. A third advantage is that, while features are \emph{domain dependent}, they are \emph{role independent}. We subsequently illustrate some intuitive semantic properties that arise from the features as a natural consequence of this.

{\bf Preliminary Results: }We describe some results obtained by using labeled data provided by the FEIII 2017 challenge organizers. The details of the challenge and dataset are described on the website\footnote{\url{https://ir.nist.gov/dsfin/2017-challenge.html}}. These results were obtained prior to the test dataset being released by challenge organizers, or test results being submitted (and scored) by participants. The original dataset contained 20 roles, of which 10 were `plural' equivalents of 10 singular roles\footnote{For example, \emph{affiliate} and \emph{affiliates}.}. For better generalization, we combined singular-plural equivalents so that there were ten unique roles. 

On a prototypical pre-challenge evaluation for three of the ten roles, namely \emph{affiliate, trustee} and \emph{issuer} across three partitions of the data (10\%, 50\% and 90\% across each role used for training, with the rest withheld for testing), we obtained precision and recall metrics across the three roles (Table \ref{exp1}). In no case is the F1-Measure below 90\%, and in many cases it is over 95\%. In future work, we will be validating these results further across roles and with more experimental trials; however, early results do seem to indicate that a promising word embedding-based approach is a promising avenue for obtaining accurate role relevance predictions without investing significantly in manual feature engineering effort.

The \emph{post-challenge} evaluation results were largely consistent with the numbers in Table 1. For example, against a (withheld) ground-truth where highly relevant triples had to be ranked over relevant triples, which in turn had to be ranked over neutral and irrelevant triples (in that order), to receive optimal scores, our approach scored 91.62\% NDCG \cite{dsmmreport}. 

{\bf Semantic Similarity Experiment: }Even more interesting are the semantics captured by the word embeddings. To show this qualitatively, we located the nearest neighbors of three `seed' keywords, namely `company', `regulators' and `city', in the word embedding vector space (Table \ref{exp2}). Results show that the embedding space has captured the `meaning' of the words remarkably well in a contextual sense, despite not having much data (incl. training+working) per word embedding standards. This is in agreement with previous results obtained over large corpora, but to the best of our knowledge, has never been satisfactorily illustrated for smaller domain-specific corpora such as in the finance/SEC domain. Nonetheless, we do believe that with more data, the results should further improve. 

{\bf Future Work:} The approach and results described in this paper were highly preliminary, as to the best of our knowledge, word embeddings have thus far not been used for financial role relevance prediction. We believe that many avenues for future work exist; at present, we are exploring (1) intelligent ways of computing CFVs rather than just normalized average (e.g., using recently proposed techniques such as paragraph2vec), (2) using external information, including distant supervision, to boost training power, (3) pooling labeled data between classifiers to minimize labeling effort.   
\begin{table}[t]
\centering
\caption{Preliminary empirical results for three roles. The header indicates the amount of training data used (the rest used for testing). Metrics are Precision/Recall percentages.}
\begin{tabular}{|p{0.9cm}|p{2.0cm}|p{2.0cm}|p{2.0cm}|} \hline
{\bf Role}&{\bf 10}&{\bf 50}&{\bf 90} \\ \hline
affiliate & 93.71/95.51 & 93.10/95.29  & 94.12/94.12 \\ \hline
trustee & 95.37/99.20  & 93.58/99.03  & 97.67/100.00  \\ \hline
issuer & 90.48/95.00 & 98.15/89.83  & 90.91/90.91  \\ \hline

\end{tabular}
\label{exp1}
\end{table} 

\begin{table}[t]
\centering
\caption{Qualitative semantic similarity results.}
\begin{tabular}{|p{2.4cm}|p{5.1cm}|} \hline
{\bf Seed keyword}&{\bf Top 3 Nearest Neighbors} \\ \hline
{company} & {bankamerica, hedging, j.p} \\ \hline
{regulators} & {monitor, organizations, supervised} \\ \hline
{city} & {representative, national, na} \\ \hline
\end{tabular}
\label{exp2}
\end{table} 

\bibliographystyle{abbrv}
\bibliography{sigproc}  

\end{document}